%% file: main.tex
\documentclass[journal]{IEEEtran}
\usepackage{graphicx,booktabs,multirow,amsmath,amssymb,array,url,xcolor,siunitx,balance,microtype,placeins,capt-of,cuted}
\usepackage[nocompress]{cite}
\usepackage[hidelinks]{hyperref}
\newcommand{\framework}{RetiDual-Select}
\newcommand{\kermany}{Kermany-4}
\newcommand{\octc}{OCT-C8}
\newcommand{\octdl}{OCTDL}
\newcommand{\pp}{percentage points}
\newcommand{\specialist}{RETFound}
\newcommand{\generalist}{MedSigLIP}
\title{RetiDual-Select: Reliability-Aware Specialist--Generalist Routing for Retinal OCT under Image-Quality Shift}
\input{author_metadata.tex}
\begin{document}
\raggedbottom
\setlength{\stripsep}{7pt plus 2pt minus 2pt}
\maketitle
\begin{abstract}
Adaptive fusion can repair a static ensemble's errors, but can also overturn correct decisions. \framework{} combines retinal specialist \specialist{} with the image encoder of the medical vision--language model \generalist{}, separating candidate fusion, routing safety and predictive review. Calibrated product-of-experts (PoE) fusion is the static reference, a small gate proposes an alternative for each image, a validation-frozen harm guard decides whether that candidate may replace the reference, and a separate score prioritizes review. Static PoE reached the highest or joint-highest clean accuracy on all three datasets. At a clean 2\% accepted-error target the review policy accepted 94.3\% of \kermany{} images at 99.89\% selective accuracy, and 92.4\% of \octc{} at 98.65\%. Both branches were correct on 95.0\% and 91.8\% of clean images; only 4.4\% and 6.2\% had exactly one correct branch. Across 15 prespecified synthetic conditions, adaptive gains were concentrated in severe contrast reduction, where this routing opportunity widened roughly ninefold; percentile rescaling largely removed the effect, linking it to compressed dynamic range. In post-hoc validation-only analyses, gain was higher when the gate followed the more reliable branch, and a frozen guard cut the worst-condition \octc{} loss from 1.39 to 0.36 \pp{}. Adaptive routing was useful when image quality created a reliability opportunity and the gate followed it; otherwise the system retained Static PoE or deferred for review. The findings are bounded by the synthetic perturbations evaluated and do not establish robustness to natural acquisition or scanner shift.
\end{abstract}

\section{Introduction}
\IEEEPARstart{O}{ptical} coherence tomography (OCT) \cite{huang1991oct} provides cross-sectional views of retinal microstructure and is central to macular-disease assessment. Deep learning has enabled accurate OCT classification and referral support \cite{kermany2018cell,defauw2018natmed}, while retinal and medical foundation models now provide reusable visual representations \cite{zhou2023retfound,sellergren2025medgemma}. High average accuracy does not show whether a model stays dependable when image appearance changes, or whether an input-dependent ensemble knows when to trust one branch over another.

This is especially relevant for heterogeneous foundation models, meaning large pretrained encoders reused for a downstream task with a small task-specific classifier. A retinal-domain specialist and a broadly pretrained medical encoder may respond differently to contrast, blur, noise and compression. Static PoE combines both branches without selecting a winner, whereas an adaptive gate can exploit a branch-specific advantage but can also replace a correct PoE decision with an error. The practical question is when adaptive fusion may depart from a strong static reference.

Trustworthy-imaging guidance emphasizes reliability under distribution change, transparent failure analysis and human oversight rather than accuracy alone \cite{kondylakis2026futureai,panayides2026medicalai}, and uncertainty-aware retinal foundation-model studies likewise motivate confidence estimation and review-oriented decisions \cite{peng2025fmue,wang2025uavlfm}. Less attention has been paid to the safety of image-specific fusion itself: when does routing repair PoE, when does it harm PoE, and can that harm be identified without tuning a policy per test condition?

\begin{figure*}[t]
\centering
\includegraphics[width=\textwidth]{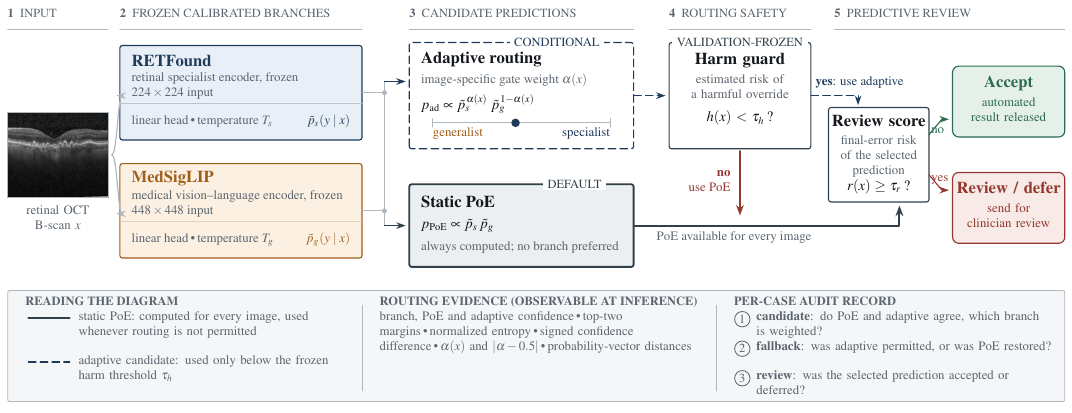}
\caption{\framework{} trustworthy-routing workflow. Frozen specialist and generalist image encoders produce calibrated class probabilities. Static PoE and image-specific adaptive routing are candidate predictions. A separate validation-frozen harm guard uses observable routing evidence to permit adaptive fusion or return the decision to PoE; a secondary risk pathway can support review/defer.}
\label{fig:pipeline}
\end{figure*}

We study these questions using \specialist{} as a retinal specialist and the visual encoder of \generalist{} as a medical generalist. \framework{} separates three decisions that are often conflated (Fig.~\ref{fig:pipeline}): PoE and the adaptive gate produce candidate class distributions, a harm guard decides whether the adaptive candidate should replace PoE, and a separate risk pathway identifies selected predictions that warrant review. This keeps routing safety distinct from predictive uncertainty and makes each fallback observable; the review pathway accepts over 92\% of clean images at 98.65\% to 99.89\% selective accuracy.

The contributions are:
\begin{itemize}
\item a decision architecture that separates candidate fusion, routing safety, and predictive review, with static PoE retained as the conservative reference;
\item a calibrated specialist--generalist framework with a transparent low-capacity gate;
\item an opportunity--alignment analysis that explains both adaptive repairs and harmful routing, conducted as a post-hoc validation-only mechanism audit; and
\item a validation-frozen harm guard that turns routing-specific risk into an explicit PoE fallback and quantifies the safety--utility trade-off, evaluated as a post-hoc validation-only extension.
\end{itemize}
\section{Related Work}
\subsection{Retinal Foundation Models and Medical Generalists}
Large OCT datasets established automated disease recognition \cite{kermany2018cell}, extending classifiers built on hand-crafted descriptors, convolutional ensembles, and uncertainty-driven weakly supervised training \cite{srinivasan2014oct,rasti2018mcme,rong2019surrogate,wang2020udmil}; retinal self-supervised pretraining later supported transfer across ocular tasks \cite{zhou2023retfound}. The encoders reused for such transfer are vision transformers \cite{dosovitskiy2021vit} pretrained by masked image modelling \cite{he2022mae} or image--text contrastive objectives \cite{radford2021clip}, and vision--language pretraining provides broader visual representations \cite{zhai2023siglip}, including medically adapted encoders \cite{sellergren2025medgemma}; the same recipe motivates proposals for generalist medical models \cite{moor2023gmai}. Uncertainty-aware retinal foundation models have combined representation learning with evidential uncertainty and review-oriented decisions \cite{peng2025fmue,wang2025uavlfm}. Our focus is different: the encoders are frozen, and the object of study is the reliability of their probability-level routing.

\subsection{Adaptive Fusion and Trustworthy Selective Decisions}
PoE \cite{hinton2002poe} and deep ensembles \cite{lakshminarayanan2017deepensembles} combine predictions without image-specific routing; fixed averaging is the simplest such comparator. Mixture-of-experts methods instead learn input-dependent expert weights \cite{jacobs1991adaptive,shazeer2017moe}, including convolutional mixtures for macular OCT \cite{rasti2018mcme}. Selective prediction descends from the error--reject trade-off \cite{chow1970reject}; later variants rank predictions by softmax confidence \cite{hendrycks2017baseline}, evidential uncertainty \cite{sensoy2018edl,xia2024uncertainty}, or a learned reject option \cite{geifman2017selective}, and calibration determines whether confidence reflects empirical correctness \cite{guo2017calibration}. Learning-to-defer methods extend the reject option by modelling the human the classifier defers to \cite{mozannar2020defer}; the review pathway here instead ranks the residual risk of an already-selected prediction. A separate line of work instead adapts the model on the test distribution itself \cite{wang2021tent}, whereas here every component stays frozen and only the decision changes. Uncertainty nevertheless deteriorates under shift \cite{ovadia2019uncertainty}, reported accuracy often fails to reproduce across sites and acquisition settings \cite{zech2018variable,finlayson2021clinician}, and methodological reviews attribute part of that gap to evaluation practice rather than model capacity \cite{varoquaux2022failures}. Common-corruption benchmarks enable controlled appearance changes \cite{hendrycks2019corruptions,disalvo2024medmnistc}; they support mechanism audits but do not replace external scanner or clinical validation.

\section{Materials and Methods}
\subsection{Study Design and Evidence Boundary}
This retrospective benchmark study used training and validation data to fit branch heads, temperatures, static fusion, the primary gate, and clean reliability models. Clean and 15-condition test predictions were then frozen. The contrast controls, full-suite alignment audit, and harm guard were follow-up analyses motivated by the original condition pattern. All follow-up models and thresholds used validation data only: held-out test labels were never used for fitting, and no policy was tuned to an individual test corruption.

\subsection{Datasets and Integrity Audit}
\input{tables/tab1_datasets.tex}
Table~\ref{tab:datasets} summarizes the datasets and resampling units. The cleaned \kermany{} split contains 49,263 training, 4,990 validation, and 968 official test images \cite{kermany2018cell}. Its classes are choroidal neovascularization (CNV), diabetic macular edema (DME), drusen, and normal. The official test set was retained; training or validation images sharing a patient/case identifier with it were removed, and the final validation split was constructed at patient level. Patient/case identifiers also defined the bootstrap clusters. The cleaned \octc{} split contains 18,264 training, 2,795 validation, and 2,797 test images \cite{naren2021octc8}. Native patient identifiers are unavailable for most \octc{} images, so complete patient-level independence cannot be established; filename and exact image-content audits were performed and uncertainty was estimated by image-level resampling. Cross-dataset matching to \kermany{} later recovered identifiers for a subset and revealed patient collisions; the collision-excluded analysis is reported only as a sensitivity analysis. The \octdl{} dataset contains 2,064 images across seven classes and was evaluated using patient-level nested five-fold out-of-fold predictions \cite{kulyabin2024octdl}. The patient-overlap sensitivity findings and their limitations are summarized in Section~\ref{sec:limitations}.

\subsection{Calibrated Specialist--Generalist Framework}
Both branches first produce calibrated class probabilities. Calibration adjusts how sharp or cautious a probability distribution is; it does not change the ordering of the class logits. PoE combines the two calibrated distributions without choosing a preferred branch. The adaptive gate instead changes their relative contribution for each image. The harm guard does not predict the diagnosis. It only decides whether the adaptive candidate is sufficiently safe to replace PoE.

\specialist{} receives $224\times224$ images and \generalist{} receives $448\times448$ images. Both encoders are frozen. A linear head, meaning a single learned classification layer, maps each frozen feature vector to class logits, which are unnormalized class scores. Let $x$ denote an OCT B-scan (one cross-sectional OCT image); $b\in\{s,g\}$ denote the specialist ($s$) or generalist ($g$) branch; $y$ denote a candidate class; $C$ denote the number of classes; and $z_{b,y}(x)$ denote the logit for class $y$. The index $c$ runs over all classes. A validation-selected temperature $T_b>0$ rescales the logits of branch $b$. Its calibrated probability is
\begin{equation}
\tilde p_b(y\mid x)=\frac{\exp[z_{b,y}(x)/T_b]}{\sum_{c=1}^{C}\exp[z_{b,c}(x)/T_b]}.
\end{equation}
The product-of-experts distribution is
\begin{equation}
p_{\mathrm{PoE}}(y\mid x)=\frac{\tilde p_s(y\mid x)\tilde p_g(y\mid x)}{\sum_{c=1}^{C}\tilde p_s(c\mid x)\tilde p_g(c\mid x)}.
\end{equation}
The numerator rewards classes supported by both branches, and the denominator makes the probabilities sum to one. Equal averaging, a validation-selected static weight, and confidence arbitration were additional comparators.

\subsection{Adaptive Gate and Secondary Reliability Pathway}
The gate estimates a specialist weight $\alpha(x)\in[0,1]$. A value of one uses only the specialist, a value of zero uses only the generalist, and intermediate values combine them. Branch confidence is the largest class probability. The top-two margin is the gap between the largest and second-largest probabilities, and normalized entropy is a zero-to-one summary of distributional uncertainty. Adaptive Basic uses these quantities and the signed specialist-minus-generalist confidence difference; Adaptive Full adds distances between the complete class-probability vectors and indicators of branch disagreement.

The standardized features enter an $L_2$-regularized logistic regression. The $L_2$ penalty discourages unnecessarily large coefficients, while $C_{\mathrm{gate}}=0.30$ is its inverse regularization strength. The gate is trained to identify which branch assigns greater probability to the true class. Five-fold out-of-fold validation means every validation prediction comes from a fold model that did not use that sample. Adaptive probabilities are a weighted geometric combination:
\begin{equation}
\begin{aligned}
\log p_{\mathrm{ad}}(y\mid x)&=\alpha(x)\log\tilde p_s(y\mid x)\\
&\quad+[1-\alpha(x)]\log\tilde p_g(y\mid x)-\log Z(x).
\end{aligned}
\end{equation}
Here, $p_{\mathrm{ad}}$ is the adaptive class distribution and $Z(x)$ is the normalization term that makes its probabilities sum to one.

A separate pathway ranks the error risk of the final selected prediction using confidence-only, disagreement-only, or combined evidence. A 2\% target means the validation threshold aims for at most 2\% errors among accepted samples; the thresholds for 1\%, 2\% and 5\% were applied unchanged to the clean tests and every quality condition. They support empirical review decisions rather than the distribution-free guarantees that split-conformal methods provide under exchangeability with a held-out calibration set \cite{angelopoulos2021conformal}.

\subsection{Image-Quality Suite and Contrast Controls}
Clinical acquisitions vary in quality through speckle, defocus, media opacity and motion or mirror artefacts \cite{chhablani2014artifacts}, and automated OCT quality assessment is an established task \cite{wang2019octiqa}. The suite below simulates selected appearance consequences of such variation under experimental control. Five quality families were applied at three prespecified levels: S1 was mild, S2 moderate, and S3 severe. The families were Gaussian blur ($\sigma=0.75,1.5,2.5$), additive Gaussian noise (standard deviation $0.05,0.10,0.18$), contrast factors ($0.8,0.6,0.4$), brightness factors ($0.85,0.70,0.55$), and Joint Photographic Experts Group (JPEG) compression quality ($70,40,20$). Each transformation was applied to the same red--green--blue image before branch-specific resizing and normalization. Both branches then retained their clean preprocessing pipelines. Clean-output parity first verified that the implementation exactly reproduced the stored clean probabilities; only then were 112,950 degraded image--model passes run.

Because severe contrast reduction dominated the observed gain, four full-test controls were added while all models remained frozen: (i) the contrast operator in the Pillow image-processing library; (ii) an independent luminance-mean transformation $I'=\mu+\gamma(I-\mu)$, where $I$ is an input intensity, $\mu$ is the image mean, and $\gamma$ is the contrast factor; (iii) the same explicit transformation after branch resize but before normalization; and (iv) percentile rescaling of the Pillow result. Brightness-S3 served as a negative control for generic intensity reduction. Input-tensor statistics were retained for audit. Figure~\ref{fig:alignment}A--B and Table~\ref{tab:trustworthy} summarize the contrast-control results.

\subsection{Opportunity--Alignment Analysis}
We separate two requirements for useful routing: \emph{opportunity}, the size of the quality-induced change in relative branch reliability, and \emph{alignment}, whether the gate moves toward the branch that became more reliable. For each clean/degraded pair, opportunity was the signed change in specialist-minus-generalist probability on the true class, and gate direction was the corresponding change in $\alpha(x)$; a condition was sign-aligned when both changes shared a nonzero sign.

Complementarity is the fraction of images with exactly one correct branch. Each test image falls into three groups: both correct, one correct, or both wrong. The one-correct group is the hard-selection opportunity band; a \emph{hard selector} returns one branch's top-1 prediction. It is not a formal upper bound on probability-level fusion, since a normalized product can return a class that is neither branch's top-1. One direction is exact. Temperature scaling preserves each branch's class ordering, so PoE is correct whenever both branches are. Gate-selection accuracy is the fraction of those images on which the gate favours the correct branch. A \emph{repair} occurs when Adaptive Full is correct and PoE is wrong; a \emph{harm} is the reverse. Repair minus harm equals the final accuracy difference, so it summarizes outcomes without being independent evidence of mechanism.

The primary fixed-suite bootstrap retained all 15 conditions, using patient/case-level resampling for \kermany{} and image-level resampling for \octc{}. Condition resampling, contrast-S3 removal, leave-one-family-out analyses, medians and 10\% trimmed means assessed whether the mean result was concentrated in a small subset of shifts. Across all 30 dataset--condition pairs we compared aligned and misaligned gains and the association between alignment magnitude and positive gain.

\subsection{Validation-Frozen Harm Guard}
The harm guard predicts the event $H(x)=1$, defined as PoE being correct while Adaptive Full is wrong. Its inputs are observable at inference: branch, PoE, and adaptive confidence; top-two margins; normalized entropy; signed confidence differences; $\alpha(x)$ and its distance from 0.5; and the same probability-vector distances used by the gate. A class-balanced logistic regression ($C_{\mathrm{guard}}=0.25$), meaning harm and non-harm samples receive compensating weights, outputs a harm score between zero and one. A larger score means greater estimated routing risk. It was fitted with five-fold out-of-fold validation. The threshold retained the greatest adaptive usage while capturing at least 80\% of validation harms. At test time, samples below the threshold use Adaptive Full; samples at or above it fall back to PoE.

Two guards were evaluated: a clean-only guard fitted on clean validation predictions, and a shift-aware guard fitted on up to 800 validation images per dataset under all 15 corruptions, with grouped folds. Leave-one-family-out validation tested transfer to an unseen family. The test action and threshold were frozen and never selected per condition.

\subsection{Trustworthy Inference Policy}
PoE is always computed and remains the default. Adaptive Full is used only when the harm score falls below its validation-fixed threshold, and the review pathway is applied after that choice. The guard therefore asks whether routing should replace PoE, and the risk score asks whether the selected prediction should be accepted automatically.

All confidence intervals (CIs) reuse frozen predictions and policies; a 95\% CI contains the central 95\% of bootstrap estimates. Kermany uncertainty is clustered by patient/case identifier and \octc{} by image. No model or threshold is refitted inside a bootstrap sample.

\section{Results}
\subsection{Static PoE Is an Accuracy Reference, Not a Safety Signal}
\input{tables/tab2_clean_octdl.tex}
Clean performance is reported in Table~\ref{tab:clean}. On \kermany{}, PoE reached 98.76\% accuracy versus 98.14\% for \specialist{} and 96.38\% for \generalist{}. On \octc{}, PoE reached 96.42\% versus 94.64\% and 95.24\%. Adaptive Full tied PoE on \kermany{} and was 0.11 \pp{} lower on \octc{}. On \octdl{}, PoE had the highest point estimate (91.62\%) and Adaptive Full reached 91.28\% (Table~\ref{tab:clean}). These results establish PoE, rather than either branch or the adaptive gate, as the conservative accuracy reference.

The branch-outcome decomposition shows why the clean margin for any routing rule is small. On \kermany{}, both branches were correct on 920 of 968 images (95.0\%), exactly one was correct on 43 (4.4\%), and both were wrong on 5 (0.5\%). On \octc{}, the corresponding counts were 2,569 (91.8\%), 173 (6.2\%), and 55 (2.0\%). A hard-selection oracle would therefore reach 99.48\% and 98.03\%, leaving hard-selector gaps of 0.72 and 1.61 \pp{} above PoE. PoE already correctly classifies 31--36 of the 43 exactly-one-correct \kermany{} images and 73--128 of the 173 \octc{} images, so clean differences between fusion rules are constrained by how rarely the branches disagree rather than by gate capacity. This is why the study evaluates routing by its safety behaviour rather than its mean gain. That advantage does not make PoE confidence a safety certificate. On \octc{} it produced 37 errors at confidence of at least 0.95, against five for equal averaging. This is why the final-error pathway is separate, rather than fused confidence serving as the review rule.

\begin{figure*}[t]
\centering
\includegraphics[width=\textwidth]{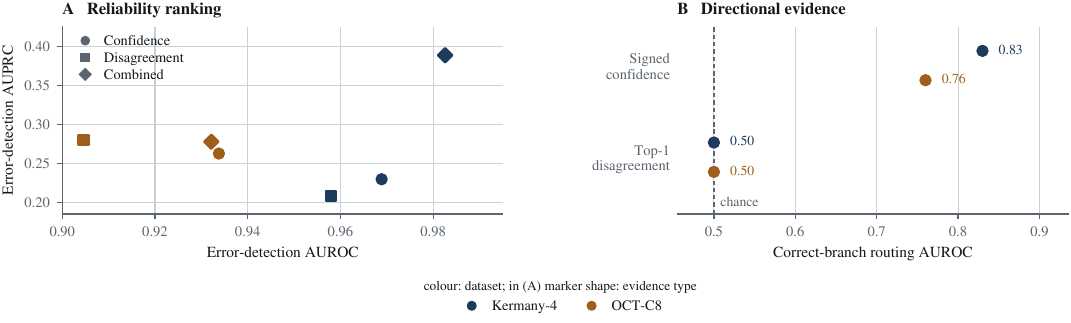}
\caption{Secondary reliability evidence on the clean internal test sets; full-coverage accuracy for every method is given in Table~\ref{tab:clean}. (A) Error-detection AUROC and AUPRC jointly show that combined evidence is strongest on Kermany-4 but not uniformly superior on OCT-C8. (B) Signed confidence contains directional routing information, whereas symmetric top-1 disagreement remains at chance.}
\label{fig:clean}
\end{figure*}
\subsection{Reliability Signals Provide Distinct Information}
The area under the receiver operating characteristic curve (AUROC) measures ranking across all decision thresholds, 0.5 being chance and 1.0 perfect; the area under the precision--recall curve (AUPRC) emphasizes the uncommon error class and therefore depends on error prevalence. Combined risk improved \kermany{} error AUROC/AUPRC from 0.969/0.230 with confidence alone to 0.983/0.389. On \octc{}, confidence retained a slightly higher AUROC (0.934 versus 0.932), while combined evidence improved AUPRC from 0.263 to 0.278. Signed confidence also ranked which branch was more likely to be correct, whereas symmetric top-1 disagreement remained at chance for routing direction. Confidence and disagreement contributed different information. Combining them was not uniformly superior across datasets (Fig.~\ref{fig:clean}).

\subsection{Clean Selective Review Identifies High-Accuracy Accepted Sets}
The clean validation-calibrated combined score produced high-accuracy accepted subsets (Table~\ref{tab:clean_selective}). At the 2\% target it reached 99.89\% selective accuracy on the 94.3\% of \kermany{} images it accepted, and 98.65\% on the 92.4\% of \octc{} images. At the 1\% target the accepted \kermany{} subset contained no errors. These are empirical operating points in the clean calibration setting only. They are not evidence of robustness. Section~\ref{sec:transfer} shows that the same thresholds do not hold their target uniformly once images are degraded.

\begin{table}[!t]
\centering
\caption{Clean validation-calibrated combined-risk operating points. Coverage is the accepted fraction; selective accuracy is accuracy within the accepted subset.}
\label{tab:clean_selective}
\footnotesize
\setlength{\tabcolsep}{2.4pt}
\renewcommand{\arraystretch}{1.03}
\begin{tabular}{@{}lrrrr@{}}
\toprule
Dataset & Target & Coverage & Errors / accepted & Selective accuracy \\
\midrule
Kermany-4 & 1\% & 83.9\% & 0 / 812 & 100.00\% \\
Kermany-4 & 2\% & 94.3\% & 1 / 913 & 99.89\% \\
OCT-C8 & 1\% & 79.9\% & 8 / 2,236 & 99.64\% \\
OCT-C8 & 2\% & 92.4\% & 35 / 2,584 & 98.65\% \\
\bottomrule
\end{tabular}
\end{table}

\subsection{Adaptive Gain Is Quality Conditional}
\input{tables/tab3_trustworthy_summary.tex}
Table~\ref{tab:trustworthy} summarizes routing relative to PoE. Across the fixed suites, Adaptive Full improved mean accuracy by 0.30 \pp{} on \kermany{} and 1.27 on \octc{}. Severe contrast reduction produced the clearest benefit (+4.65 and +19.88 points), while the median change across conditions was close to zero. Removing contrast-S3 reduced the mean difference to approximately zero on both datasets. Under severe contrast reduction, the one-correct fraction rose roughly ninefold---from 4.4\% to 42.4\% on \kermany{} and from 6.2\% to 56.6\% on \octc{}---then returned to 4.0\% and 6.1\% after percentile rescaling. Opportunity size alone did not predict condition-level gain across the 30 pairs; correct-branch selection did (Section~\ref{sec:alignment_results}). Adaptive routing therefore offered a concentrated opportunity rather than broad degradation robustness, and both the opportunity and its exploitation were separately observable.

\begin{figure*}[!t]
\centering
\includegraphics[width=\textwidth]{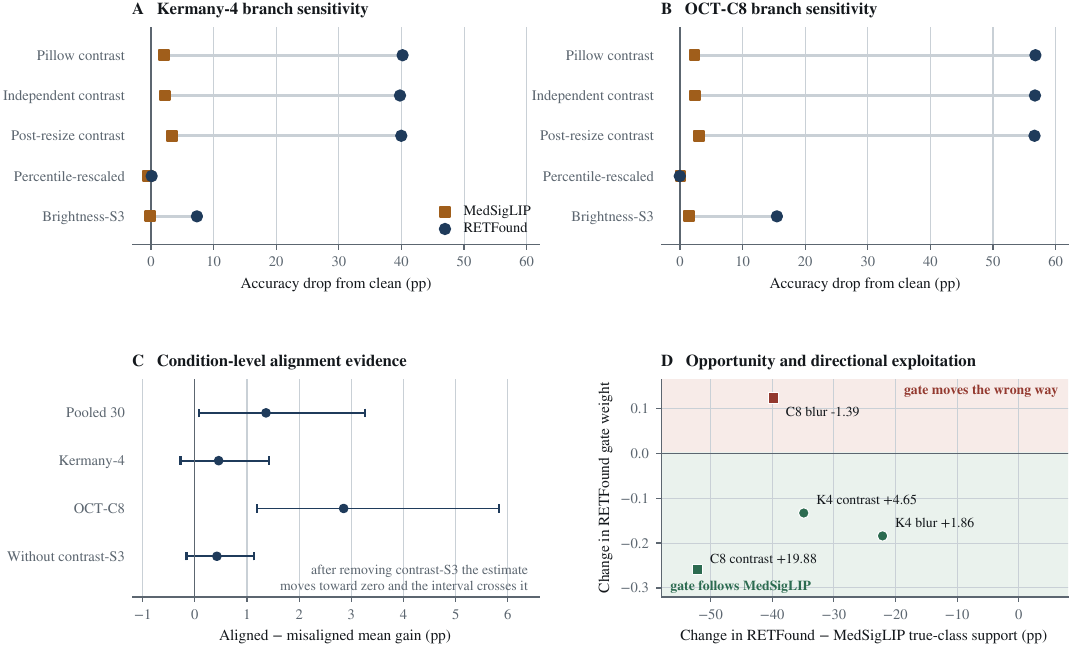}
\caption{Contrast reproducibility and opportunity--alignment audit. (A--B) Branch accuracy loss under three equivalent contrast implementations, percentile-rescaled contrast, and brightness-S3, shown on a common scale. (C) Aligned-minus-misaligned mean gain with 95\% condition-bootstrap intervals. (D) Four severe cases show why both requirements matter: contrast-S3 provides opportunity and correct alignment, whereas OCT-C8 blur provides opportunity but is misaligned and harmful. K4 and C8 denote Kermany-4 and OCT-C8; pp denotes percentage points.}
\label{fig:alignment}
\end{figure*}
\subsection{Contrast Sensitivity Is Reproducible and Reversible}
In the post-hoc validation-only contrast controls, the three equivalent implementations produced nearly identical branch behaviour (Fig.~\ref{fig:alignment}A--B). \specialist{} lost approximately 40 points on \kermany{} and 57 on \octc{}, whereas \generalist{} lost only 2--3 points. Adaptive gains remained about 5 and 19 points, respectively. Agreement across operators and preprocessing order makes a single-implementation explanation unlikely.

Percentile rescaling nearly restored clean performance and removed the adaptive advantage, linking the failure to compressed dynamic range rather than irreversible loss of anatomical content. It does not identify why the encoders differ, since architecture, resolution, normalization and pretraining also differ.

\subsection{Opportunity and Alignment Explain Success and Failure}\label{sec:alignment_results}
In the post-hoc validation-only alignment audit, aligned conditions across the 30 pairs gained more than misaligned ones (+0.964 versus -0.400 points), a difference of +1.364 (95\% condition-bootstrap CI, +0.075 to +3.264). Spearman's rank correlation was $\rho=0.437$, a positive value meaning that stronger alignment accompanied larger gain, and alignment magnitude reached AUROC 0.773 for distinguishing positive from non-positive gain. Rank-based tests did not support using alignment direction alone as a decision rule, and the difference weakened after removing contrast-S3. Alignment was informative. Useful routing also required sufficient opportunity.

The four severe cases make this distinction concrete (Fig.~\ref{fig:alignment}D). Contrast-S3 created a large shift toward \generalist{} and the gate followed, producing +4.65 and +19.88 points; Kermany blur was also aligned and gained +1.86. In OCT-C8 blur, reliability shifted toward \generalist{} but the gate moved toward \specialist{}; correct-branch selection fell to 0.348 and Adaptive Full was 1.39 points worse than PoE.

\subsection{The Harm Guard Limits Downside with a Dataset-Dependent Trade-off}
In the post-hoc validation-only guard analysis, the shift-aware guard improved both average utility and downside protection on \octc{}. Mean gain rose from +1.27 to +1.33 points and the worst condition from -1.39 to -0.36. Negative conditions fell from nine to four. It also changed blur-S3 from -1.39 to +1.00 while retaining +17.34 points under contrast-S3 (Fig.~\ref{fig:guard}).

On \kermany{}, the guard moved the operating point toward safety: the worst condition improved from -0.83 to -0.41 points, but mean gain fell from +0.30 to +0.10 and less of the contrast benefit was retained. The same fallback rule therefore produced different safety--utility balances across datasets rather than a uniform accuracy gain. Its mean and worst-condition benefits persisted across the \octc{} sensitivity subsets. The reduction in the number of negative conditions did not.

\subsection{Clean Review Targets Do Not Transfer Uniformly Under Controlled Degradation}\label{sec:transfer}
The strong clean operating points did not transfer uniformly to degraded images. At the 2\% target accepted-error rate, the clean-calibrated combined score kept the accepted error at or below 2\% in 12 of 15 \kermany{} conditions but in only two of 15 on \octc{}. Patient-level \octdl{} evaluation did not reproduce an adaptive advantage either, so adaptive fusion has not been shown to benefit an independent retinal dataset. The resulting policy is deliberately hierarchical: retain PoE under unvalidated shift, use a frozen guard only where its harm ranking has been validated, and send high-risk selected predictions to review.

\section{Discussion}
\subsection{Principal Finding: Opportunity Plus Alignment}
The central finding is conditional. Routing becomes useful when image degradation creates a meaningful branch-reliability opportunity and the gate follows it. It is not a claim of uniform adaptive superiority. Severe contrast reduction satisfied both requirements. Many mild conditions were aligned but offered too little opportunity to change the final diagnosis. On clean images the branches disagreed in outcome on only 4.4\% and 6.2\% of cases, leaving limited room for routing to improve on PoE. OCT-C8 blur showed the opposite failure: the branches separated, but the gate moved in the wrong direction.

The contrast controls strengthen this interpretation: independent operators and a preprocessing-order control reproduced the same branch asymmetry, while percentile rescaling restored near-clean behaviour. For this known synthetic transformation, rescaling is the more direct remedy. Routing is not a substitute for a validated preprocessing correction. Its narrower role is to provide a decision layer without requiring the corruption type to be identified at inference.

\begin{figure*}[!t]
\centering
\includegraphics[width=\textwidth]{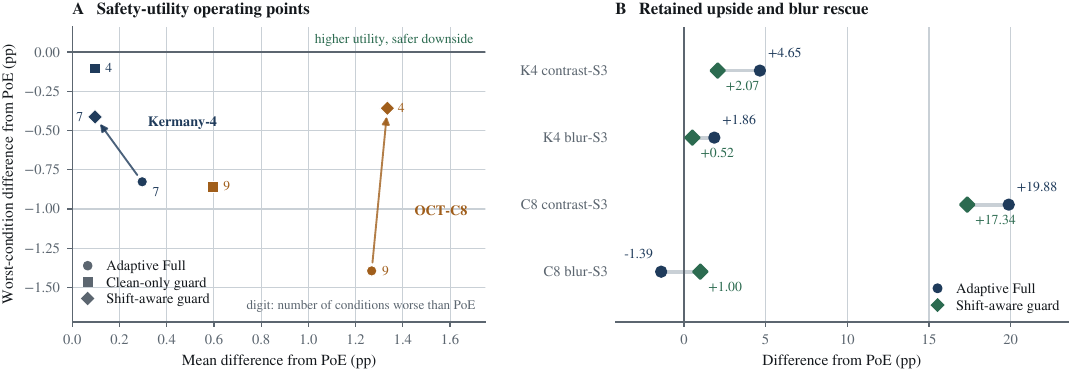}
\caption{Validation-frozen harm-guard trade-off. (A) Methods are positioned by mean utility and worst-condition downside; labels give the number of negative conditions, and arrows show the change from Adaptive Full to the shift-aware guard. (B) Contrast-S3 benefit and blur-S3 behaviour before and after fallback. The guard strongly reduces OCT-C8 downside but trades away more adaptive upside on Kermany-4. K4 and C8 denote Kermany-4 and OCT-C8; pp denotes percentage points.}
\label{fig:guard}
\end{figure*}

\subsection{Decision Hierarchy and Safety--Utility Trade-off}
Static PoE remains the anchor because it combines both branches without preferring one. It is an accuracy reference. It is not a safety certificate: its 37 high-confidence errors on \octc{} show why fused confidence cannot also serve as the review decision. Adaptive routing can exploit a large branch difference, but it introduces one specific failure: PoE is correct and the adaptive candidate is wrong. The harm guard makes this failure actionable. A low score permits the adaptive candidate. A high score returns the case to PoE, and the final-error score then decides acceptance or review.

Fig.~\ref{fig:guard} makes the trade-off explicit: on OCT-C8 the shift-aware guard moved the operating point toward both higher mean utility and smaller worst-condition loss, whereas on Kermany-4 it reduced the worst loss but also removed useful contrast repairs. The guard should be judged by mean gain, worst-condition loss and negative-condition count together. No single number represents safety.

Trustworthiness here is procedural: the static reference is exposed, repairs and harms are measured against it, fallback rules are frozen on validation data, and review absorbs residual uncertainty. PoE, temperature scaling, logistic gating and selective review are established components. The contribution is how they are organized into an auditable routing-safety protocol. Separate candidate, fallback and review records keep repairs, harmful overrides, missed repairs and deferrals distinguishable. Neither risk score is treated as a diagnosis.

\subsection{Scope, Clinical Interpretation, and Limitations}\label{sec:limitations}
\specialist{} and \generalist{} were selected before degradation testing to represent retinal-domain and broader medical pretraining. Only the \generalist{} image encoder was used; no text generation was evaluated. Replication with other OCT-capable encoders is needed before the observed branch sensitivity can be treated as a general specialist--generalist property. Confidence remained a strong error-ranking signal, but clean targets did not transfer consistently under the controlled degradations, so the framework should support human review rather than autonomous diagnosis.

Five limitations define the scope of the present evidence. First, it is retrospective and uses public B-scan benchmarks. Second, patient independence cannot be established for most of \octc{}, as Section~III-B notes. Among \octc{} test images traceable through \kermany{}, 724 correspond to patients also present in \octc{} development data. The collision-excluded re-aggregation is a sensitivity analysis, not proof of complete independence, and image-level bootstrap intervals cannot account for unknown within-patient dependence. Third, the quality suite applies synthetic perturbations to already-processed B-scans. It is a mechanism stress test, not an equivalent of naturally degraded acquisitions or of scanner, site, workflow or disease-population shift, none of which this study evaluates. Fourth, only the clean evaluation and the 15-condition suite were prespecified; the mechanism analysis and both harm guards, labelled post-hoc throughout, were not evaluated on an untouched external or prespecified shift test and therefore generate hypotheses rather than confirm them. Percentile rescaling was a sensitivity control, not a validated clinical remedy, and one encoder pair cannot separate architecture, resolution, normalization and pretraining. Fifth, a cross-dataset hash audit found 31.74\% of \octc{} images, including 885 of its 2,797 test images, to be byte-identical to \kermany{} images, so the two benchmarks partially overlap and their agreement is not independent replication. Each dataset uses its own head, temperatures, gate and guard, all fitted on its own validation data. The cross-dataset overlap therefore does not by itself create training-to-test leakage. The recovered patient collisions are a separate matter, and remain a within-dataset independence limitation. We re-aggregated the frozen per-case outputs twice, once without the shared images and once without the 724 patient-colliding images. Mean adaptive gain, contrast-S3 dominance, severe-blur failure and worst-condition guard protection were all preserved in these sensitivity analyses, while clean \octc{} accuracy increased. The reduction in negative-condition count does not transfer consistently.

A prospective study should freeze all thresholds in advance, use patient-level OCT volumes from multiple devices and centres, retain naturally low-quality acquisitions, and report harm capture, repair retention, fallback frequency, review burden and worst-subgroup performance.

\subsection{Practical Reporting, Monitoring, and Falsifiability}
The two risk scores answer different questions. The harm score estimates the risk that Adaptive Full overturns a correct PoE decision; a high score triggers fallback. The final-error score is applied afterwards to accept the selected prediction or send it for review. Both scores lie between zero and one, but both are tied to their validation procedures rather than to universal clinical probabilities.

A deployment report should keep candidate, fallback and review actions separate, stating how often each was taken with the four outcome counts defined above, since these make the safety--utility trade-off visible when accuracy alone cannot. Before prospective evaluation, preprocessing, branch heads, temperatures, the gate, both risk models and all thresholds should be frozen, with action frequencies and failure counts stratified by acquisition quality and clinical subgroup. Shifts in branch confidence, gate weight, fallback frequency and review burden can then trigger investigation, though such monitoring shows only that behaviour has changed, not why.

\section{Conclusion}
\framework{} reframes OCT foundation-model fusion as an auditable routing problem. Static PoE provides a conservative reference; adaptive weighting is most useful when image quality creates a substantial branch-reliability opportunity and the gate follows that direction. Equivalent contrast controls demonstrated a reproducible, reversible dynamic-range sensitivity, and a post-hoc validation-frozen harm guard reduced routing downside in these benchmark tests. Together, these results define an auditable decision framework for when to permit adaptive fusion, retain PoE, or request review. External device- and patient-level validation remains the next step toward clinical use, since most externally validated radiologic models show lower external performance \cite{yu2022external}.

\FloatBarrier
\input{declarations.tex}

\end{document}

%% file: author_metadata.tex
\author{Zihao Han and Philippe De Wilde%
\thanks{Zihao Han and Philippe De Wilde are with the University of Kent, Canterbury, Kent CT2 7NZ, United Kingdom.}%
\thanks{Corresponding authors: Zihao Han (e-mail: zh219@kent.ac.uk) and Philippe De Wilde (e-mail: p.dewilde@kent.ac.uk). Zihao Han is the primary contact.}%
\thanks{This work has been submitted to the IEEE for possible publication. Copyright may be transferred without notice, after which this version may no longer be accessible.}%
}

%% file: tables/tab1_datasets.tex
\begin{table*}[!t]
\centering
\caption{Datasets, split information, uncertainty units, and evaluation roles. ``Out-of-fold'' means that each reported sample was predicted by a fold model that did not use that sample for fitting. Patient identifiers were unavailable for OCT-C8, so its uncertainty estimates use image-level resampling.}
\label{tab:datasets}
\footnotesize
\setlength{\tabcolsep}{4.0pt}
\begin{tabular}{@{}llrrrll@{}}
\toprule
Dataset & Classes & Train & Validation & Test / out-of-fold & Resampling unit & Primary role \\
\midrule
Kermany-4 & 4 & 49,263 & 4,990 & 968 & patient/case & Clean evaluation and 15-condition quality suite \\
OCT-C8 & 8 & 18,264 & 2,795 & 2,797 & image & Clean evaluation and 15-condition quality suite \\
OCTDL & 7 & \multicolumn{2}{c}{nested five-fold development} & 2,064 & patient & Independent-task boundary \\
\bottomrule
\end{tabular}
\end{table*}

%% file: tables/tab2_clean_octdl.tex
\begin{table*}[!t]
\centering
\caption{Clean internal test performance and patient-level OCTDL out-of-fold performance. Values are percentages. The F1 score is the harmonic mean of precision and recall; Macro-F1 is the unweighted mean of the class-specific F1 scores. Best accuracy in each dataset is bold; ties are bolded together.}
\label{tab:clean}
\footnotesize
\setlength{\tabcolsep}{5.0pt}
\begin{tabular}{@{}lrrrrrr@{}}
\toprule
& \multicolumn{2}{c}{Kermany-4} & \multicolumn{2}{c}{OCT-C8} & \multicolumn{2}{c}{OCTDL} \\
\cmidrule(lr){2-3}\cmidrule(lr){4-5}\cmidrule(l){6-7}
Method & Accuracy & Macro-F1 & Accuracy & Macro-F1 & Accuracy & Macro-F1 \\
\midrule
RETFound & 98.14 & 98.14 & 94.64 & 94.63 & 88.86 & 80.72 \\
MedSigLIP & 96.38 & 96.41 & 95.24 & 95.24 & 87.55 & 78.54 \\
Equal average & 98.55 & 98.55 & 96.35 & 96.35 & 91.52 & 85.62 \\
Static PoE & \textbf{98.76} & 98.76 & \textbf{96.42} & 96.42 & \textbf{91.62} & 85.42 \\
Static weighted & 98.45 & 98.45 & 96.35 & 96.35 & 90.84 & 84.47 \\
Confidence arbitration & 98.55 & 98.55 & 96.39 & 96.38 & 91.28 & 85.25 \\
Adaptive Basic & 98.66 & 98.66 & 96.17 & 96.16 & 91.57 & 85.56 \\
Adaptive Full & \textbf{98.76} & 98.76 & 96.32 & 96.31 & 91.28 & 84.92 \\
\bottomrule
\end{tabular}
\end{table*}

%% file: tables/tab3_trustworthy_summary.tex
\begin{table*}[!t]
\centering
\caption{Trustworthy-routing summary relative to static PoE. The symbol $\Delta$ denotes an accuracy difference in percentage points. Equivalent contrast ranges cover Pillow, independent luminance-mean, and post-resize implementations.}
\label{tab:trustworthy}
\footnotesize
\setlength{\tabcolsep}{5.5pt}
\renewcommand{\arraystretch}{1.08}
\begin{tabular}{@{}lrrp{0.40\textwidth}@{}}
\toprule
Metric & Kermany-4 & OCT-C8 & Interpretation \\
\midrule
Fixed-suite Adaptive Full $\Delta$ & +0.30 & +1.27 & Positive mean gain, but concentrated in severe contrast reduction. \\
Mean without contrast-S3 & -0.01 & -0.06 & No evidence of broad degradation robustness. \\
RETFound contrast drop range & 39.77--40.19 & 56.63--56.78 & Reproducible specialist sensitivity across equivalent implementations. \\
MedSigLIP contrast drop range & 2.07--3.31 & 2.32--3.04 & Relative generalist stability under the same controls. \\
Adaptive Full contrast-S3 $\Delta$ & +4.65 & +19.88 & Large reliability opportunity with correct directional exploitation. \\
Adaptive Full blur-S3 $\Delta$ & +1.86 & -1.39 & Similar quality change can help or harm depending on routing direction. \\
Shift-aware guard mean $\Delta$ & +0.10 & +1.33 & Guarding trades utility for downside protection in a dataset-dependent way. \\
Worst $\Delta$: Adaptive Full $\rightarrow$ guard & -0.83 $\rightarrow$ -0.41 & -1.39 $\rightarrow$ -0.36 & Worst-condition harm is reduced on both datasets. \\
Negative conditions: Adaptive Full $\rightarrow$ guard & 7 $\rightarrow$ 7 & 9 $\rightarrow$ 4 & Fewer conditions fall below PoE, and only on OCT-C8. \\
\bottomrule
\end{tabular}
\end{table*}

%% file: declarations.tex
\section*{Declarations}
\textit{Data availability:} This study used publicly available, de-identified retinal optical coherence tomography datasets. Dataset sources, cleaned split sizes, and a summary of the integrity and overlap audits are documented in this manuscript. Original images remain subject to the source-dataset licences.

\textit{Code and reproducibility:} This preprint presents the main article, including its methods, principal results, figures, and tables. It does not include the separate supplementary information, experiment manifests, source-data files, or figure-generation scripts accompanying the journal submission. Third-party model and dataset licences continue to apply.

\textit{Ethics and competing interests:} No new participants were recruited and no personally identifiable patient information was accessed; only de-identified public-dataset case identifiers were used for split and overlap auditing. The work was limited to retrospective computational analysis of public de-identified benchmark datasets. The authors declare no competing interests related to this work.